\documentclass[journal]{IEEEtran}
\ifCLASSINFOpdf
\else
\fi

\usepackage{graphicx}
\usepackage{subfig}
\usepackage{amsmath,amssymb} 
\usepackage{color}
\usepackage{cite}
\usepackage{multirow}
\usepackage{booktabs}
\usepackage{verbatim}

\begin{document}

\newcommand{\etal}{\emph{et al}.}

%
\title{Temporally Coherent Video Harmonization Using Adversarial Networks}
%
%
%

\author{Haozhi Huang,
	Senzhe Xu,
	Junxiong Cai,
	Wei Liu,
	Shimin Hu
}
%
%

\markboth{Journal of \LaTeX\ Class Files,~Vol.~14, No.~8, August~2015}%
{Shell \MakeLowercase{\textit{et al.}}: Bare Demo of IEEEtran.cls for IEEE Journals}
%



\maketitle

\begin{abstract}
Compositing is one of the most important editing operations for images and videos. The process of improving the realism of composite results is often called harmonization. Previous approaches for harmonization mainly focus on images. In this work, we take one step further to attack the problem of video harmonization. Specifically, we train a convolutional neural network in an adversarial way, exploiting a pixel-wise disharmony discriminator to achieve more realistic harmonized results and introducing a temporal loss to increase temporal consistency between consecutive harmonized frames. Thanks to the pixel-wise disharmony discriminator, we are also able to relieve the need of input foreground masks. Since existing video datasets which have ground-truth foreground masks and optical flows are not sufficiently large, we propose a simple yet efficient method to build up a synthetic dataset supporting supervised training of the proposed adversarial network. Experiments show that training on our synthetic dataset generalizes well to the real-world composite dataset. Also, our method successfully incorporates temporal consistency during training and achieves more harmonious results than previous methods.
\end{abstract}


%
\IEEEpeerreviewmaketitle

\section{Introduction}
%
%
%
%

\begin{figure*}[t]
	\begin{center}
		\includegraphics[width=1.0\linewidth]{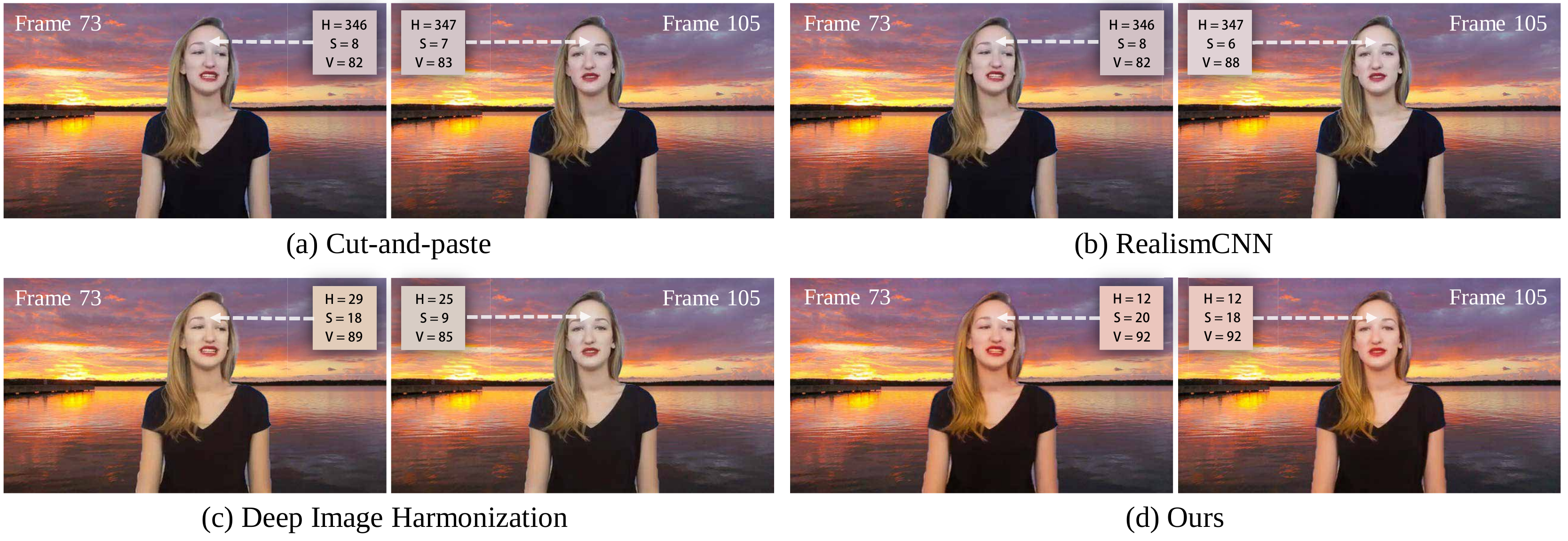}
	\end{center}
	\caption{Given a composite video generated by a direct cut-and-paste operation (a), our method learns to harmonize it to imporve its realism. Compared with previous methods  RealismCNN~\cite{zhu2015learning} (b)  and  Deep Image Harmonization~\cite{tsai2017deep} (c), the harmonized results of our method look more natural and temporally consistent (d). The warm color during sunset is correctly cast to the foreground in our result. HSV color values at the same position are shown for clear comparison.}
	\label{fig:teaser}
\end{figure*}

\IEEEPARstart{G}{enerating} realistic composite videos is a fundamental requirement in video editing tasks. Given two videos, one of them contains a desired foreground, and the other contains a desired background. To generate a realistic composite video from these two source videos, three steps are needed to be performed correctly. First, extract a foreground object from one of the source videos by computing an alpha matte or a binary mask indicating the pixels which belong to the foreground. Second, paste the foreground to a proper location in the background.  Third, adjust the foreground appearance to make it look natural in the new background. The last step is often called \textbf{harmonization}. In this paper, we focus on the video harmonization task which intends to improve the realism of a composite video by performing appearance adjustments on the foreground (see Figure~\ref{fig:teaser}). 

Traditional methods for handling image harmonization are based on learning statistical relationships between hand-crafted appearance features~\cite{cohen2006color, lalonde2007using, sunkavalli2010multi, xue2012understanding}. These methods neglect whether the foreground and the background are compatible considering the context of one whole image. Recently, some image harmonization methods are proposed to leverage  powerful convolutional neural networks (CNNs) to automatically learn features that capture context and semantic information of the composite images, which generate more appealing harmonization results. Zhu \emph{et al}.~\cite{zhu2015learning} trained a CNN to distinguish natural images from generated ones. Then they used the predicted realism score to guide a simple color adjustment of the foreground to obtain more realistic composite images.  Unlike the method in~\cite{zhu2015learning} which takes realism evaluation and improvement as two separated steps, Tsai \etal~\cite{tsai2017deep} proposed an end-to-end network which takes a composite image and a foreground mask as input to generate a harmonized image directly. 

All the methods mentioned above aim at harmonization for images. For videos, although applying image harmonization frame by frame can generate a video harmonization result, this will introduce obvious flicker artifacts in the absence of considering the temporal consistency between consecutive harmonized frames. In this paper, we propose an end-to-end harmonization network for videos, which is able to simultaneously harmonize the composite frames and maintain the temporal consistency between them. In order to make the harmonization results more realistic, we propose to train the network with a pixel-wise discriminator. Different from the most common global discriminators used in the literature which predict whether an image is real or fake as a whole, our discriminator is trained to precisely distinguish the harmonious pixels from the disharmonious ones. In addition, with the well trained pixel-wise discriminator, we are able to predict foreground masks automatically to relieve the need of input foreground masks. On the other hand, for the purpose of constraining the network to generate temporally consistent results, we train it with a temporal loss term to incorporate temporal information in the training phase and avoid the trouble of computing optical flows in the inference phase. 

Training the proposed end-to-end harmonization network requires a large amount of video data taken at different scenes with ground-truth harmonious and disharmonious pairs, foreground masks and optical flows, which is still a missing piece in the community. To this end, we propose a simple yet effective way to build up a synthetic dataset which satisfies this demand. We also build up a real-world composite video dataset for evaluating the proposed method in real scenarios, which helps demonstrate that training on our synthetic dataset enables our network to generalize to real-world composite videos. Extensive experiments on the two datasets demonstrate that the proposed method is able to conduct temporally coherent video harmonization while generating more harmonious results than existing methods.

The contributions of our method are two-fold. Firstly, to the best of our knowledge, this is the first end-to-end CNN for video harmonization. The network is trained in an adversarial way with an introduced temporal loss to simultaneously acquire high-quality harmonization results and temporal consistency. Secondly, a synthetic dataset is proposed for supporting the efficient training of the video harmonization network, which contains ground-truth harmonious and disharmonious pairs, foreground masks and optical flows.

\section{Related Work}
Our work attempts to generate a temporally consistent harmonized video by an adversarial network. This is closely related to the  literature on image harmonization, conditional generative adversarial networks, and temporally consistent video editing.

\smallskip
\noindent\textbf{Image harmonization.} Traditional methods for image harmonization focus on matching appearance statistics without considering the context of images, such as aligning the statistics of global or local histograms~\cite{reinhard2001color, xue2012understanding}, shifting the colors towards predefined harmonious color templates~\cite{cohen2006color}, gradient-domain compositing~\cite{perez2003poisson, jia2006drag, tao2010error}, multi-scale matching of various statistics~\cite{sunkavalli2010multi}, and maximizing the co-occurrence probability of color distributions~\cite{lalonde2007using}. Recently, Zhu \etal~\cite{zhu2015learning} trained a discriminative model based on a CNN to predict a realism score for a composite image, and then used the score to determine a simple brightness and contrast adjustment for improving the realism. Instead of separating realism evaluation and improvement into two steps, Tsai \etal~\cite{tsai2017deep} proposed an end-to-end CNN to learn how to harmonize a composite image directly. To improve the ability of capturing semantic information, Tsai \etal~pretrained the harmonization network with semantic segmentation and used the segmentation branch to provide features to help harmonization. Different from these existing algorithms which are engaged in image harmonization, we propose an end-to-end CNN trained in an adversarial manner to solve the problem of video harmonization, which generates realistic and temporally consistent harmonized results. 

\smallskip
\noindent\textbf{Conditional image generation based on adversarial training.}
Generative adversarial network (GAN) was first proposed by Goodfellow \etal~\cite{goodfellow2014generative} to address the problem of realistic image generation from input noise variables. The key idea of GAN is to train a generator and a discriminator in an adversarial fashion. While the discriminator is trained to distinguish fake images from real ones, the generator is trained to deceive the discriminator and generate images as realistic as possible. Recently, GANs have also been widely used in the task of conditional image generation~\cite{pathak2016context, ledig2016photo, isola2016image, zhu2017unpaired, mathieu2015deep, liang2017dual}. 
Although the Markovian discriminator proposed in~\cite{isola2016image,shrivastava2017learning} performs a patch-level discrimination instead of a global one, it is designed for speeding up the discriminator with a very small receptive field, which is still insufficient for the harmonization task. In this paper, we propose to use an encoder-decoder structure to learn a pixel-wise discriminator which labels each pixel as harmony or disharmony for calculating a more precise adversarial loss.
With the proposed discriminator, we can also complete the harmonization task without input foreground masks.

\smallskip
\noindent\textbf{Temporally consistent video editing.} Directly applying image harmonization frame by frame for videos inevitably results in flicker artifacts. This is because the corresponding regions in different frames are harmonized in different ways. Plenty of approaches have been proposed to enforce temporal consistency in video editing, such as spatio-temporal smoothing ~\cite{lang2012practical,bonneel2013example,aydin2014temporally}, optimization with a temporal loss~\cite{bonneel2015blind,ruder2016artistic},  frame propagation~\cite{ye2014intrinsic}, {\em etc}. The above methods either require extra post-processing operations or rely on a time-consuming optimization process. 
Recently, some video style transfer methods \cite{gupta2017characterizing,huang2017real} show that temporal consistency and style transfer can be simultaneously learned by a CNN, which acquires considerable temporal consistency at a very little time cost. Inspired by this idea, our proposed method also chooses to incorporate the temporal consistency during the training phase. However, instead of calculating a global temporal loss, we introduce a regional temporal loss that forces our model to pay more attention to the disharmonious regions, therefore leading to more coherent harmonized results.

\begin{figure*}[t]
	\begin{center}
		\includegraphics[width=1.0\linewidth]{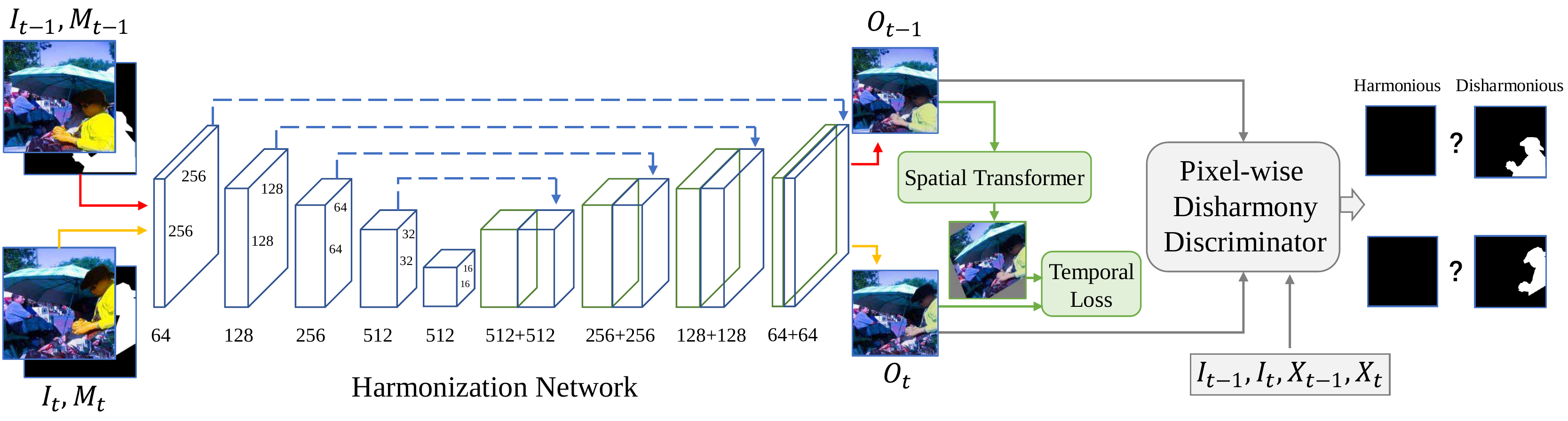}
	\end{center}

	\caption{The training phase of the proposed video harmonization model. The harmonization network is trained in an adversarial manner with the pixel-wise disharmony discriminator. A two-frame coordinated training strategy is adopted to incorporate a regional temporal loss to constrain the consecutive harmonized foregrounds to have similar appearances.}
	\label{fig:model}

\end{figure*}

\begin{figure}[t]
	\centering
	\resizebox{1.0\linewidth}{!}{
		\centerline{
			\subfloat[Ground-truth frame1]{\includegraphics[width=0.33\linewidth]{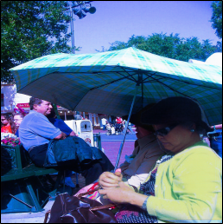}} \hspace{0.1pt}
			\subfloat[Foreground mask]{\includegraphics[width=0.33\linewidth]{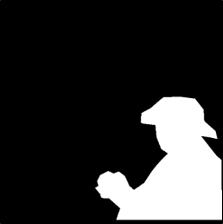}}
			\hspace{0.1pt}
			\subfloat[Pure background]{\includegraphics[width=0.33\linewidth]{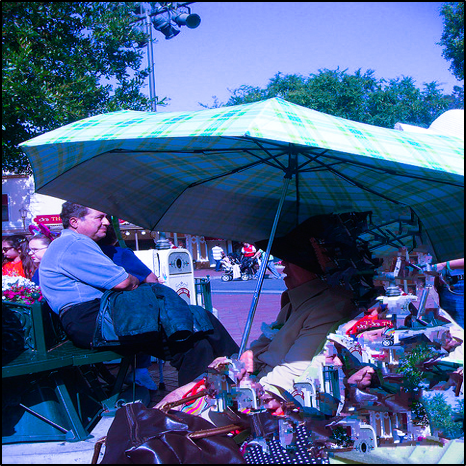}}
		}
	}
	\resizebox{1.0\linewidth}{!}{
		\centerline{
			\subfloat[Composite frame1]{\includegraphics[width=0.33\linewidth]{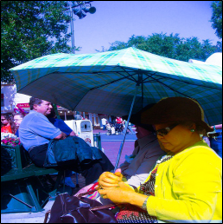}}
			\hspace{0.1pt}  
			\subfloat[Ground-truth frame2]{\includegraphics[width=0.33\linewidth]{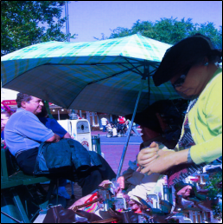}} 
			\hspace{0.1pt}
			\subfloat[Composite frame2]{\includegraphics[width=0.33\linewidth]{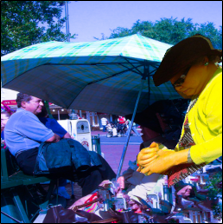}} 
		}
	}

	\caption{Building up the synthetic dataset. Given an image (a), we take it as the first ground-truth frame. Then we cut out the foreground and apply inpainting to obtain the pure background (c). By performing color adjustment on the foreground of (a), we obtain the first composite frame (d). By applying a random affine transform to the foregrounds of (a) and (d), we obtain the second ground-truth frame (e) and the second composite frame (f).}
	\label{fig:dataset}

\end{figure}

\section{Video Harmonization Network}

In this section, we describe the details of our proposed end-to-end CNN for video harmonization. Figure~\ref{fig:model} shows an overview of our network. The harmonization network takes one frame of a composite video and a foreground mask as input, and performs appearance adjustments on the foreground while keeping the background unchanged. To incorporate temporal consistency between consecutive harmonized frames, a two-frame coordinated training strategy with a regional temporal loss is adopted. Note that in the training phase the two frames are fed to the harmonization network in a coordinate but separate way, while in the testing phase the harmonization network processes a video in a frame by fame way. This kind of setting has been proven to be effective for training a network which conducts smoother transformation with less flicker artifacts in video style transfer~\cite{huang2017real}. To further enhance the realism of the harmonized results, the harmonization network is trained in an adversarial way with a pixel-wise disharmony discriminator, which distinguishes the disharmonious pixels from the harmonious ones. Moreover, the well-trained discriminator can also be employed to predict the disharmony area in the input, which holds as a replacement of the input foreground mask.

\subsection{Synthetic Training Dataset}\label{sec:synthetic}

Before delving deeply into the network architecture, it is essential to describe the way we collect data. For supervised training, our harmonization network needs a composite video and a corresponding harmonized video as a sample pair. Given an arbitrary composite video, it is hard to acquire a high-quality harmonized result even for a human expert. For the image harmonization task, Tsai \etal~\cite{tsai2017deep} collected images from the MSCOCO dataset~\cite{lin2014microsoft} which have ground-truth foreground masks, and then applied color transfer between random foreground pairs with the same semantic labels. While the image after a foreground adjustment is used as the input, the original image is used as the ground-truth. The difficulty of extending this idea to the video harmonization task lies in the fact that there are a limited number of videos that have ground-truth foreground masks. Even in the very recent video object segmentation dataset DAVIS~\cite{Pont-Tuset_arXiv_2017}, there are only 90 annotated videos. This number is far from being enough, because the harmonization network requires training data covering tremendous scenes to learn the natural appearances of foregrounds in various cases. Meanwhile, we also need the ground-truth optical flows between consecutive frames for evaluating the temporal consistency.

To address this data issue, we construct a synthetic dataset named Dancing MSCOCO which contains ground-truth foreground masks and optical flows. Based on the MSCOCO dataset, we apply small-scale random affine transforms to the foregrounds and acquire a series of images containing the same ``dancing" foreground, which simulate the consecutive frames in a real video. A similar strategy has been adopted by Dosovitskiy\etal~\cite{dosovitskiy2015flownet} to collect data for training a FlowNet to predict optical flows, which shows competitive performance compared to state-of-the-art methods like DeepFlow~\cite{weinzaepfel2013deepflow} and EpicFlow~\cite{revaud2015epicflow}. Similarly, Khoreva \etal \cite{khoreva2017lucid} used synthetic frames to help the training of an object tracking system. The success of \cite{dosovitskiy2015flownet, IMKDB17, khoreva2017lucid} proves that proper synthetic data is sufficient to some extent for training a deep neural network targeting at dealing with real video data.

Key outputs during the process of building up our Dancing MSCOCO dataset are shown in Figure~\ref{fig:dataset}. First, we select images containing 'people' as the foregrounds from the MSCOCO dataset, and wipe off the images whose foreground area is smaller than $10\%$ of the whole image. Since the image numbers of different classes are imbalanced in the MSCOCO dataset, training all the classes together inevitably introduces biases. In this paper, we simply focus on images containing people to avoid a class bias. It is easy to transfer the proposed method to other kinds of foregrounds. Further solutions for avoiding a class bias are beyond the scope of this paper. Second, we cut out the foreground and apply inpainting~\cite{criminisi2004region} to fill the holes to obtain pure background images. For each background image, we perform random cropping to obtain a distorted copy of the background image and resize it back to the original size, which simulates a background movement in a video. Third, we apply color adjustments to the foregrounds to simulate the composite images. Besides performing color transfer~\cite{reinhard2001color} between random foreground pairs as in~\cite{tsai2017deep}, we also perform random adjustments of the basic color properties including exposure, hue, saturation, temperature, contrast, and tone curve. This makes our dataset cover more kinds of compositing situations than the dataset created in~\cite{tsai2017deep}. Fourth, we apply the same random affine transform to the original foreground and the color adjusted foreground, which simulates a foreground movement in a video, and paste it back to the corresponding randomly cropped background. Since we know the exact affine transform between the foregrounds, it is easy to acquire the corresponding ground-truth optical flow.  The affine transform parameters are sampled from a suitable range in order not to conduct large-scale foreground movements. Finally, we achieve a total number of 33,338 pairs of "consecutive original frames" and their corresponding "consecutive composite frames". 29,818 pairs of them are used as the training data, 1,000 pairs are used as the validation data, and 2,520 pairs are used as the testing data. We have also tried generating more than one distorted copy for each image, but found a marginal improvement in the training.


\subsection{Adversarial Training with Temporal Loss}

Our network contains two parts, a harmonization network $G$ behaving as a generator and a pixel-wise disharmony discriminator $D$. The harmonization network $G$ processes a composite video frame by frame to generate a harmonized video. At each time step $t$, $G$ takes a composite frame $I_t$ and a foreground mask $M_t$ as input and generates a harmonized output frame $ O_t= G(I_t, M_t)$, which adjusts the appearance of the foreground to make it look more natural in the background. To incorporate temporal consistency, the network $G$ is trained in a two-frame coordinated manner to constrain the consecutive outputs $O_{t-1}$ and $O_{t}$ to be temporally coherent. Thanks to this training strategy, $G$ is able to generate temporally consistent harmonized frames, though processing a video in a frame-by-frame way. The insight behind this strategy is that by constraining the outputs to be temporally consistent, we train a more stable network that can avoid amplifying small differences in the inputs. This strategy makes the harmonization network learn smoother transformations. To acquire more realistic harmonized results, we propose a pixel-wise disharmony discriminator $D$ to play against $G$ by telling disharmonious pixels from harmonious ones. 

\smallskip
\noindent\textbf{Pixel-wise disharmony discriminator.}
The objective for training $D$ is to classify harmonious pixels into class $0$ and disharmonious ones into class $1$, while keeping the parameters of $G$ fixed. The loss function we aim to minimize is:
\begin{equation} \small
\begin{aligned}
\mathcal{L}_{D} = &\frac{1}{2N} \big\|D(O_t)-M_t \big\|_2^2 + \frac{1}{2N} \big\|D(I_t)-M_t \big\|_2^2  \\ &+ \frac{1}{N} \big\|D(X_t) \big\|_2^2.
\end{aligned}
\end{equation}
Here, $M_t$ is the input foreground mask, in which the foreground pixels are labeled as $1$ and the background pixels are labeled as $0$. $D(O_t)$ is the discriminator output for $O_t$, which should be close to $M_t$ for guiding the discriminator to distinguish the harmonized pixels from the ground-truth realistic pixels. In our experiments, we find that training $D$ merely with $O_t$ cannot generalize well to $I_t$, which is also a kind of image with a disharmonious foreground. Thus, we also feed $I_t$ to $D$ as a disharmonious sample, and give the loss terms relevant to $O_t$ and $I_t$ the same weight. $D(X_t)$ is the discriminator output for the ground-truth realistic frame $X_t$, in which all pixels should be labeled as $0$. Note that, unlike the most common global discriminators in the literature which consider an image as a whole to be real or fake, our discriminator learns to classify each pixel separately. This is because the background pixels in $O_t$ and $I_t$ are definitely harmonious, which should be treated differently from those disharmonious pixels for training a more precise discriminator. To correctly label a pixel, the discriminator should have a large receptive field to capture context information. Thus a UNet~\cite{ronneberger2015u} architecture is used as the harmonization network, which will be described in detail later. Theoretically, our method is a variant of LSGAN~\cite{mao2016least}, which shows a more stable and faster convergence than vanilla GAN~\cite{goodfellow2014generative} or WGAN~\cite{arjovsky2017wasserstein}. 

\smallskip
\noindent\textbf{Harmonization network.}
The objective for training $G$ is to generate temporally coherent harmonized frames that are indistinguishable from realistic ones. The loss function that we minimize for training $G$ is:
\begin{equation} \small
\label{eq:harmonization}
\begin{aligned}
\mathcal{L}_{G} = & \underbrace{ \frac{1}{N} \big\|O_t-X_t \big\|_2^2}_{\text{Reconstruction Loss}}
+ \underbrace{ \frac{\lambda_1}{N_F} \left\|M_t \circ \big(O_t - S(O_{t-1}\big)) \right\|_2^2  }_{\text{Regional Temporal Loss}} 
\\ & + \underbrace{\frac{\lambda_2}{N} \big\|D(O_t) \big\|_2^2}_{\text{Adversarial Loss}}.
\end{aligned}
\end{equation}
Here, $N$ denotes the total number of pixels in a frame, and $N_F$ denotes the number of pixels in the foreground. The harmonization network is trained with a combination of a reconstruction loss, a regional temporal loss, and an adversarial loss. The reconstruction loss enforces the harmonized frame $O_t$ to be similar to the ground-truth realistic frame $X_t$. The temporal loss enforces the harmonized frame $O_t$ to be temporally consistent with the previous harmonized frame $O_{t-1}$. $S$ denotes a Spatial Transformer Network~\cite{jaderberg2015spatial}, which warps $O_{t-1}$ according to the ground-truth optical flow provided by our Dancing MSCOCO dataset. The Spatial Transformer Network is fully differentiable, which makes our network end-to-end trainable. Instead of treating each pixel equally as in the global temporal loss~\cite{huang2017real}, our regional temporal loss focuses on the foreground region by taking a Hadamard product of the foreground mask and the global element-wise difference between frames: $M_t \circ \big(O_t - S(O_{t-1}\big))$. The insight behind this is that since the network only needs to learn a simple  identity mapping for the background pixels, it should pay more attention to learning how to generate temporally consistent foregrounds. In our experiments, we find that the generated backgrounds achieve great temporal consistency even without any applied temporal loss. 

On the other hand, the adversarial loss encourages $G$ to generate harmonized results that are indistinguishable for the discriminator $D$, hence enforcing the output of $G$ to be close to the manifold underlying realistic frames. Although the reconstruction loss has enforced the harmonized frames to be similar to the realistic frames, this is not enough because there may be multiple answers besides the ground-truth for harmonizing a composite frame. Different training samples with similar compositing situations may teach $G$ to learn different harmonization solutions for similar inputs. This finally prompts $G$ to output an average of different solutions, which may fall outside the manifold of realistic frames.
Leveraging $D$ to provide an adversarial loss for training $G$ can deal with this problem.

As shown in Figure~\ref{fig:model}, our harmonization network adopts the architecture like UNet~\cite{ronneberger2015u}, which has skip connections to reserve more content details which may be lost during the progressive downsampling in the convolutional layers. Our discriminator also adopts the same architecture for acquiring a large receptive field. The network proposed for deep image harmonization~\cite{tsai2017deep} also used skip connections but in the form of element-wise summation instead of feature concatenation in the UNet. We have also tried the residual network used in style transfer~\cite{johnson2016perceptual}. The superiority of the UNet over other networks is clearly validated in Section~\ref{sec:evaluate_network}.

\subsection{Harmonization without Foreground Masks}

Previous state-of-the-art harmonization methods all require a foreground mask as the input besides the composite image itself. In this paper, with the help of a well-trained pixel-wise disharmony discriminator, we can predict the disharmonious foreground area automatically. Thus, we are able to accomplish the harmonization task without an input foreground mask using the same harmonization network which is trained with foreground masks: $O_t = G\big(I_t, D(I_t)\big)$.
Another solution for achieving the same goal is to train $G$ without input foreground masks in the first place. In our experiments, we find that this solution achieves similar performance to ours but comes with an extra labor of training another network. 

\section{Experiments}

In this section, we present extensive experiments to evaluate the effectiveness of the proposed method.

\smallskip
\noindent\textbf{Real-world composite dataset.} Besides the synthetic dataset, we also build up a dataset containing real-world composite videos to evaluate the effectiveness of the proposed method. First, we collect videos with the tag 'fashion' from Youtube-8M dataset~\cite{abu2016youtube}, the content of which usually is a person talking in front of a static camera. We require the videos to be taken by a static camera because the incompatible movements of cameras also influence the realism of a composite video, which is out of the scope of this paper. For each video, we cut out a 5-10 seconds clip which contains the desired content. To generate the ground-truth foreground masks, we utilize the Rotobrush tool in Adobe After Effects CC 2017 to manually label the foreground regions. Additionally, we collect various background videos from Videvo.net~\cite{videvo}. For each background video, we apply random adjustments of basic color properties as in Section~\ref{sec:synthetic} to get two more distorted copies to cover various composite situations. In total, we acquire 30 foreground videos and 48 background videos. Then we extract the foregrounds paste them to different backgrounds one by one. In the end, we get 1440 composite videos. For this real-world composite dataset, the ground-truth optical flows between frames are unavailable. Instead, we utilize the state-of-the-art method~\cite{IMKDB17} to estimate the optical flows.

\smallskip
\noindent\textbf{Implementation details.} 
During training, we resize the input and the ground-truth frame to $512 \times 512$, and scale the range of color values to $[-1, 1]$. We set $\lambda_1=2 \times 10^{-2}$ and $\lambda_2=10^{-2}$ with a fixed learning rate of $2 \times 10^{-4}$. We use a batch size of $1$ to alternatively train our harmonization network and pixel-wise disharmony discriminator for $45$ epochs. Here, the batch size $1$ means for each iteration we feed only one pair of consecutive frames for training. For optimization we use Adam~\cite{kingma2014adam} with $\beta_1=0.999$.  Hyper parameters are chosen by experiments on the validation dataset.

\begin{table*}[t]
	
	\begin{center}

		\caption{Comparison with previous methods. Quantitative results are evaluated on the synthetic dataset and the real-world composite dataset. The user study results show the Plackett-Luce scores acquired by different methods }
		\label{tab:cmp_previous}
		\resizebox{0.7\textwidth}{!}{
		\begin{tabular}{|c|c|c|c|c|c|c|c|}
			\hline
			\multirow{2}{*}{Method} &
			\multicolumn{4}{c|}{Quantitative Results} &
			\multicolumn{2}{c|}{User Study Results} \\
			\cline{2-7}
			&  PSNR & MSE  & $\mathcal{L_\text{T1}}$ &  $\mathcal{L_\text{T2}}$ &  Realism & Temporal  \\
			\hline
			Cut-and-paste & 18.19 & 0.030 & \textbf{0.0006}  & 0.027 & -0.004  & \textbf{2.176}  \\
			Zhu~\cite{zhu2015learning} & 18.27 & 0.032 & 0.0822 & 0.098 & -0.330 & -2.064\\
			Tsai~\cite{tsai2017deep} & 18.42 & 0.024 & 0.2325 & 0.029 & -0.179 & -1.291\\
			Ours & \textbf{22.45} & \textbf{0.009}  & 0.0247 & \textbf{0.026} & \textbf{0.551} & 1.179\\
			\hline
		\end{tabular}
		}
	\end{center}
\end{table*}

\begin{figure*}[t]
	\begin{center}
		\centering
		\resizebox{1.0\linewidth}{!}{
			\includegraphics[width=1.0\linewidth]{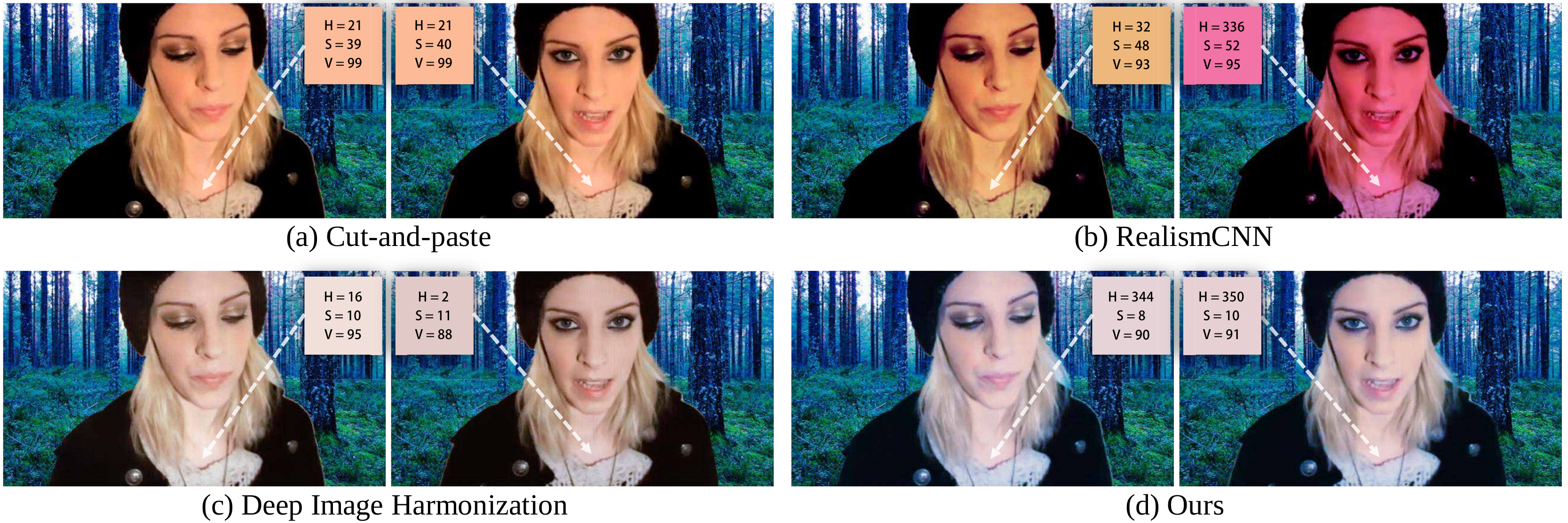}		
		}
	\end{center}%

	\caption{Visual results on the real-world composite dataset. （(a) are the cut-and-paste results, {\em i.e.}, the inputs to the harmonization methods. (b) are the results of RealismCNN~\cite{zhu2015learning}. (c) are the results of Deep Image Harmonization~\cite{tsai2017deep}. (d) are our results. Among all methods, our harmonized results appear most harmonious and temporally consistent. The foregrounds of our results are adjusted to cool color tone which looks more harmonious with the background. }
	\label{fig:result1}

\end{figure*}

\begin{figure*}[t]
	\begin{center}
		\centering
		\resizebox{1.0\linewidth}{!}{
			\includegraphics[width=1.0\linewidth]{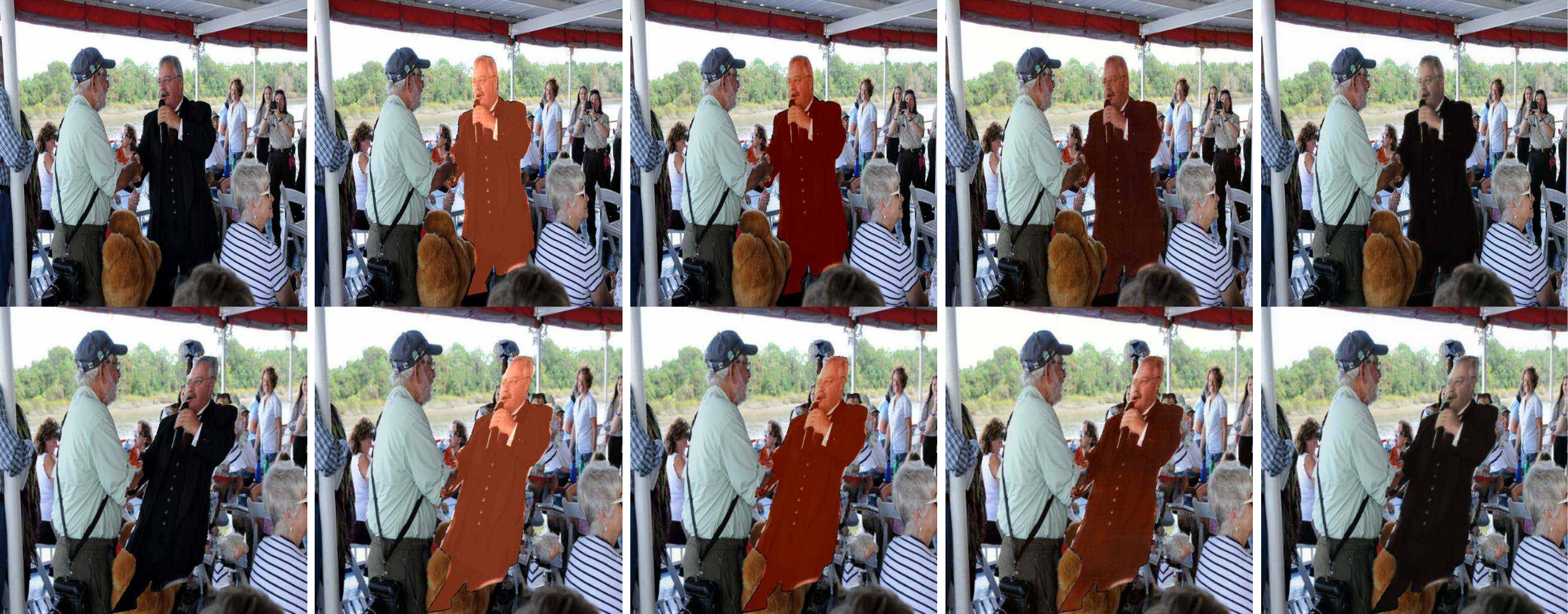}
		}
	\end{center}%

	\caption{Visual results on the synthetic dataset. The first column are the ground-truth harmonious frames. The second column are the input frames. The third column are the results of RealismCNN~\cite{zhu2015learning}. The fourth column are the results of Deep Image Harmonization~\cite{tsai2017deep}. The last column are our results. Among all methods, our harmonized results are closest to the ground-truths and most temporally coherent.}
	\label{fig:result2}

\end{figure*}

\subsection{Comparison with Previous Methods}\label{sec:cmp_previous}

We evaluate the proposed method through comparisons with two state-of-the-art methods~\cite{zhu2015learning,tsai2017deep} and the cut-and-paste baseline. Here, we choose to compare with previous methods using their original settings and training data to demonstrate the temporal inconsistency of image harmonization methods and the effectiveness of the proposed synthetic dataset. 

Table~\ref{tab:cmp_previous} shows the quantitative evaluation results. PSNR and MSE are calculated on the synthetic dataset. $\mathcal{L_\text{T1}}$ represents the regional temporal loss computed according to Equation~\ref{eq:harmonization} using the ground-truth optical flow in our synthetic dataset.  $\mathcal{L_\text{T2}}$ represents the regional temporal loss computed using the estimated optical flow in the real-world composite dataset. We show that our method achieves better performance compared to the state-of-the-art methods~\cite{zhu2015learning,tsai2017deep} in terms of both PSNR, MSE and temporal losses. Although the cut-and-paste baseline shows better temporal consistency, its realism is far from being satisfactory. 

To demonstrate training on our synthetic dataset is able to generalize well to real-world composite videos. We set up a user study similar as~\cite{zhu2015learning,tsai2017deep} using 15 real-world composite videos randomly picked from the real-world composite dataset, in which each user watches four videos and is asked to rank the harmonized results of the four methods regarding to either single frame realism or temporal consistency between frames. As a result, a total of 32 subjects participate in this study with a total of 480 rankings over the four candidate methods. Then we use the Plackett-Luce (P-L) model~\cite{maystre2015fast} to compute the global ranking score for each method. Table~\ref{tab:cmp_previous} shows that compared with the other harmonization methods, our method achieves the highest ranking score according to both single frame realism and temporal consistency. It is no surprise that the cut-and-paste method achieves the best temporal consistency score, but it comes at the cost of realism because no appearance adjustment is applied.

Figure~\ref{fig:teaser}, Figure~\ref{fig:result1} and Figure~\ref{fig:result2} illustrate the visual results generated by different methods. Overall, our method generates more harmonious and temporally consistent results than previous methods. RealismCNN~\cite{zhu2015learning} may generate unsatisfactory results when the realism prediction process fails. Among all the methods, the foreground colors of our results are the most consistent with the backgrounds. In addition, both \cite{zhu2015learning} and \cite{tsai2017deep} generate foregrounds in different appearances across frames, which leads to flicker artifacts in the video. More visual comparisons can be found in the supplementary material.

\subsection{Effectiveness of The Synthetic Dataset}\label{sec:evaluate_dataset}

To evaluate the effectiveness of the synthetic dataset, we train our harmonization network without the discriminator using different datasets, and test on the synthetic dataset and real-world composite dataset. The datasets we have tested include the MSCOCO dataset~\cite{lin2014microsoft}, the DAVIS dataset~\cite{Pont-Tuset_arXiv_2017}, and our Dancing MSCOCO dataset. For the MSCOCO dataset, similar as the preprocessing for our synthetic dataset, we only keep those images containing people as the foregrounds and those with foreground occupying over $10\%$ of the whole image, and apply color transfers between foregrounds to acquire composite images as in~\cite{tsai2017deep}. For the DAVIS dataset, we apply random adjustments to the basic color properties to create $62$ recolored copies for each of the original 70 videos, resulting in 4410 videos in the end. When training on the DAVIS dataset, we use the same temporal loss setting as in our proposed model. We also train on our dataset without the temporal loss for comparison.
Table~\ref{tab:cmp_dataset} shows that training on our dataset with our regional temporal loss $\mathcal{L}_T$ achieves smaller MSE and temporal losses than training on the other datasets. This means that our Dancing MSCOCO dataset generalizes the model to more composite cases, while the temporal loss incorporated during the training phase leads to temporally consistent harmonized results in the inference phase. Although training on the DAVIS dataset leads to a smaller temporal loss on the real-world composite dataset, it comes at the sacrifice of the single frame realism. This is because DAVIS dataset covers limited number of scenarios, which makes generalizing to other dataset very difficult.

\begin{table}[t]
	\begin{center}
		\caption{Performances of training on different datasets.}
		\label{tab:cmp_dataset}
		\resizebox{0.45\textwidth}{!}{
		\begin{tabular}{ccccc}
			\toprule
			Dataset & PSNR & MSE  & $\mathcal{L_\text{T1}}$ &  $\mathcal{L_\text{T2}}$\\
			\midrule
			MSCOCO & 21.38 & 0.011 & 0.051  & 0.027 \\
			DAVIS & 17.11 & 0.031 & 0.060 & \textbf{0.015}\\
			Ours (w/o $\mathcal{L_\text{T}}$) & \textbf{23.29} & \textbf{0.008} & 0.043 & 0.033 \\
			Ours & 22.13 & 0.009 & \textbf{0.022} & 0.023 \\
			\bottomrule
		\end{tabular}
	}
	\end{center}

\end{table}

\subsection{Analysis of The Network Architecture}\label{sec:evaluate_network}

\noindent\textbf{Model choice.} To justify the choice of the UNet architecture, we train different networks without the discriminator and using the same setting on the Dancing MSCOCO dataset. The network architectures we have tested include the UNet~\cite{ronneberger2015u}, the deep image harmonization network (DIH) proposed in~\cite{tsai2017deep}, the style transfer residual network proposed in~\cite{johnson2016perceptual}. The number of parameters in different architectures are constrained to the same level. Table~\ref{tab:cmp_network} shows that the UNet achieves the smallest MSE. This is because the skip connections used in the UNet reserve image details well and keep extra information that is needed by the harmonization task. Although DIH obtains the smallest $L_T1$, its MSE is the largest and PSNR is the lowest. As ensuring the harmonization quality is the first priority, we choose UNet as our network structure.

\begin{table}[t]
	\begin{center}
		\caption{Performances of different network architectures.}
		\label{tab:cmp_network}
		\resizebox{0.45\textwidth}{!}{
		\begin{tabular}{ccccc}
			\toprule
			Model & PSNR & MSE  & $\mathcal{L_\text{T1}}$ &  $\mathcal{L_\text{T2}}$\\
			\midrule
			ResNet~\cite{johnson2016perceptual} & 21.38 & 0.011 & 0.020 & 0.026 \\
			DIH~\cite{tsai2017deep} & 19.82 & 0.014 & \textbf{0.014} & \textbf{0.022}\\
			UNet~\cite{ronneberger2015u} & \textbf{22.13} & \textbf{0.009} & 0.022 & 0.023\\
			\bottomrule
		\end{tabular}
	}
	\end{center}

\end{table}

\noindent\textbf{Regional temporal loss v.s. global temporal loss.} To demonstrate the effectiveness of the proposed regional temporal loss, we train our harmonization network without the discriminator using either the regional temporal loss or the global temporal loss.  After  converging to similar  PSNRs and MSEs, we stop the training and compare the temporal losses in the foreground regions. While using regional temporal loss results in  $\mathcal{L_\text{T1}}=0.022,  \mathcal{L_\text{T2}}=0.023$, using global temporal loss results in $\mathcal{L_\text{T1}}=0.038,  \mathcal{L_\text{T2}}=0.025$. It shows that using regional temporal loss can enforce the model to create more temporally consistent foregrounds.

\begin{figure*}[t!]
	\begin{center}
		\centering
		\resizebox{1.0\linewidth}{!}{
			\includegraphics[width=1.0\linewidth]{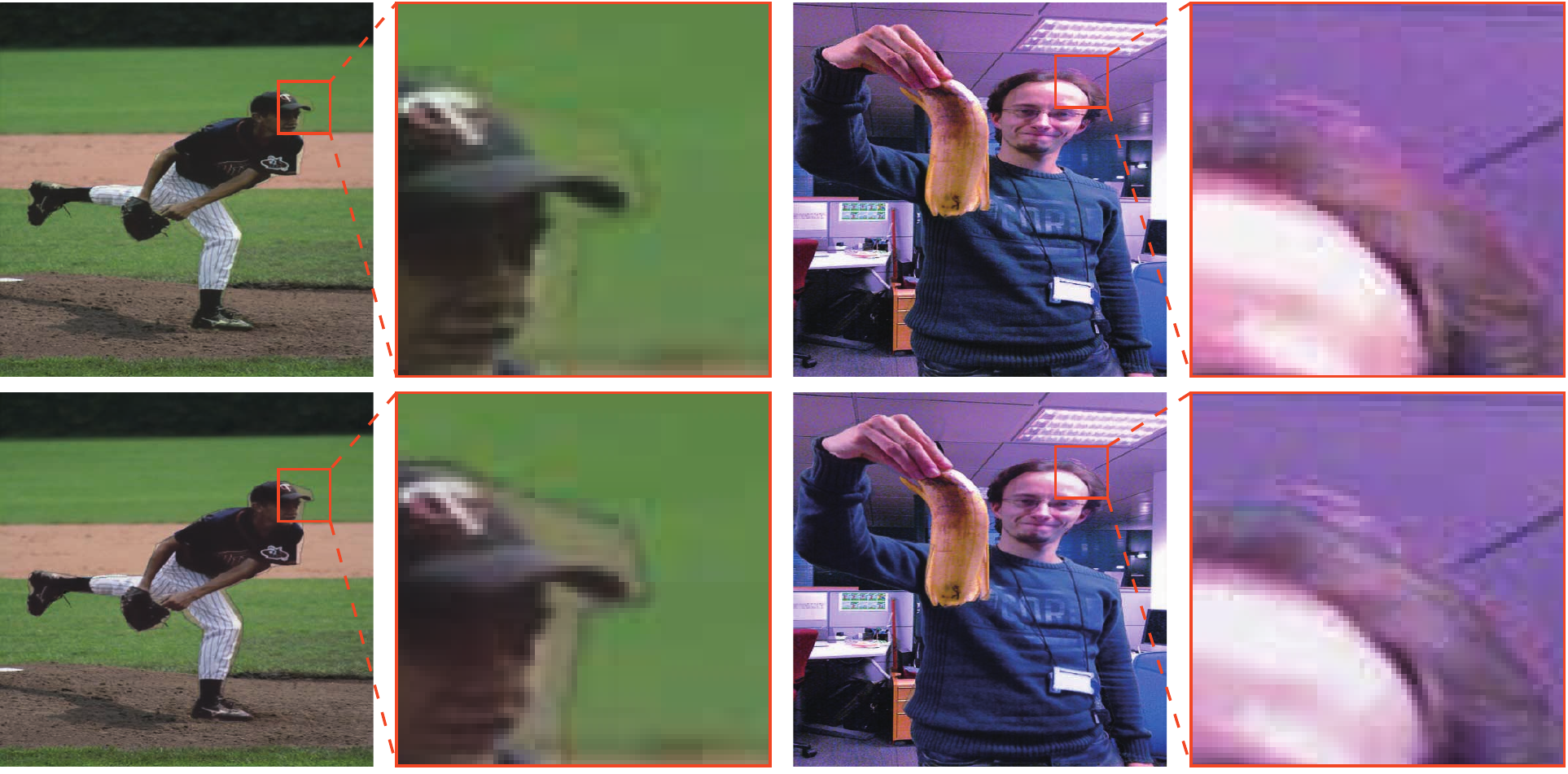}
		}
	\end{center}%

	\caption{The first row shows the results generated by the model trained with a pixel-wise disharmony discriminator. The second row shows the results generated by the model trained without a discriminator, in which the foreground boundaries are more obvious.}
	\label{fig:seggan}

\end{figure*}

\noindent\textbf{Ablation study on the adversarial loss.}
To evaluate the effectiveness of the adversarial loss, besides the proposed model, we also train a model without the discriminator. Figure~\ref{fig:seggan} shows harmonized results generated by the two models, from which we can see that the model trained with a discriminator produces harmonized foregrounds with more realistic appearances and the foreground boundaries are less obvious. We also set up a user study using 15 videos randomly picked from the real-world composite dataset, in which each user watches a pair of videos generated by the two methods at a time, and is asked to choose the one that looks more realistic. A total of 36 subjects participate in this study with a total of 540 pairwise comparisons over the two candidate methods. While $53\%$ of the choices prefer the model with the discriminator, $38\%$ of the choices prefer the one without the discriminator. The rest $9\%$ think that they are equally good. This demonstrates that the pixel-wise disharmony discriminator contributes to the generation of more realistic harmonized results.

\subsection{Evaluation of Estimated Foreground Masks}\label{sec:evaluate_mask_free}

\begin{figure}[t]
	\begin{center}
		\begin{minipage}[t]{1.0\linewidth} 
			\includegraphics[width=1.0\linewidth]{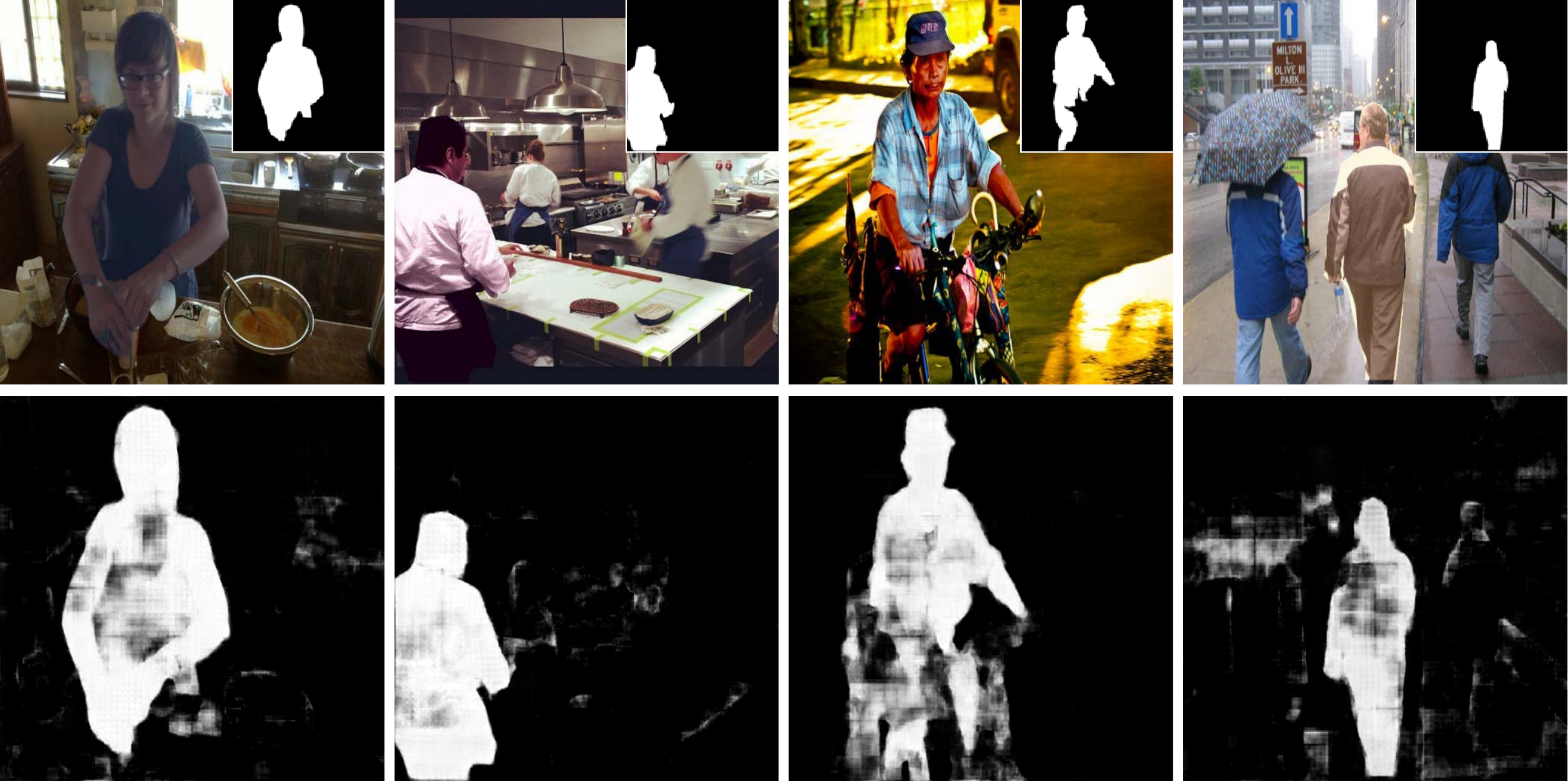}
		\end{minipage}
	\end{center}
	\caption{Disharmony maps predicted by our pixel-wise discriminator. The first row are the inputs with the ground-truth foreground masks at the top-right corner. The second row are the predicted disharmony maps.}
	\label{fig:disharmony_map}
\end{figure}

With a well-trained pixel-wise disharmony discriminator, we are able to use the predicted disharmony map instead of the ground-truth foreground mask as input to generate harmonized results. Figure~\ref{fig:disharmony_map} shows that the disharmony maps predicted by our discriminator are close to the ground-truth foreground mask and are considerable to be used as a replacement of the input foreground masks. For comparison, we test alternative solutions including using an all-zero mask or an all-one mask with our trained harmonization network, and training a new harmonization network from scratch taking no foreground mask as input.
Table~\ref{tab:mask_free} shows that while all-zero mask and all-one mask solutions result in large MSEs and low PSNRs, using a predicted disharmony map (Adversarial $D$) achieves similar performance as training a new network from scratch. This demonstrates that the discriminator endow the proposed network with the flexibility of being used either with or without input foreground masks. 

\begin{table}[t]
	\begin{center}
		\caption{Results of different solutions without input foreground masks.}
		\label{tab:mask_free}
		\resizebox{0.4\textwidth}{!}{
		\begin{tabular}{ccccc}
			\toprule
			Solution & PSNR & MSE \\
			\midrule
			All-Zero Mask & 17.72 & 0.031  \\
			All-One Mask & 14.00 & 0.048\\
			Training from Scratch & 20.04 & 0.016\\
			Adversarial $D$  & 19.87 & 0.018\\
			Full Model with Mask & \textbf{22.45} & \textbf{0.009}\\
			\bottomrule
		\end{tabular}
	}
	\end{center}
\end{table}

\section{Conclusions}
In this paper, we proposed an end-to-end CNN for tackling video harmonization. The proposed network contains two parts: a generator ( i.e., the harmonization network) and a pixel-wise discriminator. While the harmonization network takes a composite video as input and outputs a harmonized video that looks more realistic, the pixel-wise discriminator plays against it by learning to distinguish disharmonious foreground pixels from those harmonious ones. To maintain temporal consistency between consecutive harmonized frames, a regional temporal loss was adopted to enforce the harmonization network to pay more attention to generating temporally coherent harmonized foregrounds. To address the problem of lacking suitable video training data, a synthetic dataset Dancing MSCOCO was constructed for achieving harmonization and temporal correspondence ground-truths at the same time. With the help of a pretrained pixel-wise disharmony discriminator, we can relieve the necessity of an input foreground mask for gaining a considerable harmonization quality. The extensive experiments demonstrate the superiority of our method over the state-of-the-arts.


%

\section*{Acknowledgment}

The authors would like to thank Tencent AI Lab and Tsinghua University - Tencent Joint Lab for the support.

\ifCLASSOPTIONcaptionsoff
  \newpage
\fi



%
\bibliographystyle{IEEEtran}
\bibliography{IEEEabrv,mybib}

%










\end{document}